# Artificial Neural Network and Its Application Research Progress in Distillation

Jing Sun, Qi Tang

School of Chemical Engineering, East China University of Science and Technology, Shanghai 200237, China

**Abstract:** Artificial neural networks learn various rules and algorithms to form different ways of processing information, and have been widely used in various chemical processes. Among them, with the development of rectification technology, its production scale continues to expand, and its calculation requirements are also more stringent, because the artificial neural network has the advantages of self-learning, associative storage and high-speed search for optimized solutions, it can make high-precision simulation predictions for rectification operations, so it is widely used in the chemical field of rectification. This article gives a basic overview of artificial neural networks, and introduces the application research of artificial neural networks in distillation at home and abroad.

## 1 Introduction

As the research hotspot of artificial neural networks in the field of artificial intelligence since the 1980s, the application of artificial neural networks in the field of chemical engineering has become more and more extensive [1,2]. Distillation, as a typical chemical technology, is gradually becoming mature and the scale of production is constantly expanding. Therefore, the calculation requirements for rectification are becoming more and more stringent, and artificial neural networks can simulate and predict rectification operations, and the precision is high, so the application of artificial neural network in distillation has been continuously and deeply studied [3,4]. This article reviews the basic concepts of artificial neural networks and the application status and research progress of artificial neural networks in distillation.

## 2 Overview of Artificial Neural Networks

### 2.1 The structure and characteristics of artificial neural networks

The artificial neural network originated from the MP model established by the famous psychologist McCulloch and mathematical logician Pitts in 1943 [5]. Artificial neural network is a parallel dynamics system with a topology distributed structure composed of neurons and neuron interconnections, which is artificially established. It consists of an input unit that receives external signals and data, an output unit that outputs system processing results, and a hidden unit that cannot be observed outside the system between the input unit and the output unit [6]. The parallel distributed signal processing mechanism adopted by the artificial neural network makes it have the advantages of fast processing speed and strong fault tolerance. Because the artificial neural network simulates the human brain neuron network, the neuron has two states of activation and inhibition, so the mathematical relationship it represents is also non-linear, and the threshold value of the neuron can also improve the fault tolerance and storage capacity of the artificial neural network. And because a neural network usually contains multiple interconnected neurons, the interaction between the entire network and the neurons is closely related, which also makes the artificial neural network have non-limiting characteristics. The artificial neural network can handle all kinds of changing information, and the dynamic system itself is constantly changing, which also causes the artificial neural network to be very qualitative. In addition, the artificial neural network is also non-convex, that is, the dynamic system may have multiple stable equilibrium states, and the evolution is diversified. This is because the function may have multiple extreme values. The above four characteristics are also the basic characteristics of artificial neural networks.

The unique structure and information processing method of ANN makes it have obvious advantages in many aspects and has a wide range of applications. The main application areas are image processing [7-10], robot control, automatic control of power systems [11-13], signal processing [14-16], intelligent driving [17,18], health care and medical treatment[19-21], game theory [22,23], process control and optimization[24-27], etc.

In addition, in the fiber optic communication system, the artificial neural network is also used in the receiving end of the fiber optic communication system because of its characteristic learning ability, so as to learn the distortion signal of the input network, classify and identify the original sent information, and eliminate the influence of the dispersion, nonlinear and other effects in the process of signal transmission.

The artificial neural network uses computer technology to abstract the human brain neuron network and establish a calculation model, so the mathematical algorithm is the core of the artificial neural network [28].

## 2.2 Typical learning algorithms in artificial neural networks

### 2.2.1 BP algorithm

Error back propagation algorithm, referred to as BP algorithm, was proposed by Rumelhart, Hinton and Williams in 1986 [29]. BP algorithm includes two learning processes of signal forward propagation and error back propagation. It is an error correction learning algorithm. Its working principle is that after the information is input in the input layer, it is processed layer by layer through the hidden layer and passed to the output layer. If the expected output value is not reached, the output error will be propagated back to the input layer in some form, and the error signal of each layer unit will be obtained after the error is apportioned, and this will be used as the basis for correcting the weight of each unit. Keep correcting the weight until the error is allowed or the number of learning and training reaches the preset value. The artificial neural network trained according to the BP algorithm is called the BP neural network. The network has strong nonlinear mapping ability, and also has a certain fault tolerance and anti-interference ability. It is a relatively mature artificial neural network, so it is widely used [30,31].

For example, in terms of project cost, there are many factors influencing the project cost, which also determines the complexity of the project pricing basis.

However, due to the lack of accurate data in the early stage of the project, the workload will be too large if the site is measured in the later stage of the project. It is very difficult to calculate the amount of bulk materials, and the amount of bulk materials lacks accuracy due to project schedule restrictions or technical reasons, which will affect the accuracy of project cost and the scientific nature of project management. If using traditional engineering budget law, production capacity index method, the key spot check accounting method and other methods to estimate or accounting, there is a lack of consideration of the properties of the specific construction project, resulting in low estimation accuracy. Therefore, using the BP neural network that can approximate any continuous function with arbitrary precision and has good effects in nonlinear modeling, using similar engineering data to estimate the amount of bulk materials can provide more reliable data for the estimation or accounting of bulk material quantity in project cost management [32]. In addition, in terms of quality control, there are three common product quality control methods: 1. Sampling inspection method, which inspects randomly selected products, rejects waste products, and retains qualified products; 2. Statistical process control method to obtain product quality status through statistics and analysis of product quality characteristic data and give early warning to the production process; 3. Intelligent control method, establish an accurate and reasonable prediction model and predict the quality of the product, to provide a reference for optimizing the production process. Among the three methods, the first and second methods have the problem of delay in quality control; although the third method can predict the quality parameters in advance, it cannot realize the automatic adjustment of the processing process. In order to solve the above problems, a researcher has proposed a product quality prediction-control model based on BP network, which realizes the adjustment of processing parameters and optimization of processing technology by predicting the dimensional error of the workpiece, and finally achieves the purpose of effectively controlling product quality, thereby providing manufacturing companies to optimize the processing process, reduce production costs, and enhance the competitiveness of enterprises [33].

### 2.2.2  L-M method

The Levenberg-Marquardt method, referred to as the L-M method, was proposed in 1963 for a paper by D.W. Marquardt on the development of K.Levenbevg. The L-M method integrates the steepest descent method and Taylor series, and improves the shortcomings of the two. It is an optimization algorithm belonging to the nonlinear least squares algorithm, which can obtain the numerical solution of the local minimum of the nonlinearity of the function. The operation process of the L-M algorithm is to use the model function to make a linear approximation in the field of the estimated parameters, and turn it into a linear least squares problem by ignoring the derivative terms of the second order or more. The L-M algorithm has a faster convergence rate and has a wide range of applications [34].

In the micro-seismic positioning, some researchers use the L-M inversion algorithm to improve the conventional diffraction migration superposition positioning method. According to the three-component geophone embedded on the ground, the azimuth angle of the ray incident is obtained, and the location of the seismic source is searched for by the method of ray path backtracking. In the case of ensuring the calculation accuracy, the calculation efficiency is greatly improved [35]. In addition, the wax deposition rate, which is an important process parameter in the pipeline transportation of waxy crude oil, is very important for the economic operation of oil pipelines. In the traditional method of solving the wax deposition rate, the stepwise regression analysis method can only describe the linear relationship, so the solution accuracy of this method is low, but the calculation speed is faster than the artificial neural network method; the artificial neural network method has high solution accuracy, but the calculation speed is slow, and the explicit expression of the wax deposition rate model cannot be obtained. Therefore, the researchers proposed a fast and accurate Levenberg-Marquardt optimization method to search for the optimal regression parameters of the wax deposition rate-the relationship between the shear stress at the inner wall of the pipe, the temperature gradient, the dynamic viscosity of the crude oil, and the wax molecule concentration gradient, which provides a more accurate and reliable wax deposition rate mathematical model for the operation and management of oil pipelines [36].

### 2.2.3 RBF method

The radial basis function method, referred to as the RBF method, was proposed by Powell in 1985 [37]. The radial basis function is usually defined as a monotonic function of the Euclidean distance from any point in the space to a certain center. It is used to solve the problem of multivariate interpolation and is currently one of the main fields of numerical analysis research. In 1988, Moody and Darken proposed the RBF neural network using the radial basis function as the activation function [38]. The hidden layer space of the neural network is composed of RBF as the basis of the hidden unit, so the input vector can be directly mapped to the hidden layer space without weight connection. RBF neural network can avoid local minima problems, and the learning speed is faster, and it is widely used in various fields [39,40].

For example, in the actual infrared temperature measurement process, many times the emissivity of the measured target is unknown. At this time, the RBF neural network can help infrared multi-spectral temperature measurement when the emissivity of the target is not clear. Firstly, derive the strong nonlinear relationship between the target temperature and the peak radiance and its wavelength, clarify the input variables of the neural network, and then learn the multiple sets of radiance curves of the measurement target at different temperatures based on the RBF network. The temperature measurement model is established by using the mapping relationship between the wavelength at the peak point of the radiance curve, the brightness value and the temperature. This method has high measurement accuracy, and even if the peak point of the spectral radiance curve is in the water vapor and carbon dioxide absorption band or oscillation band, it is also applicable [41]. In addition, the RBF network also plays an important role in the prediction and visualization of urban waste production. With the overall increase in the level of citizen consumption, the increasing amount of garbage discharge has made the phenomenon of "garbage besieged city" a global trend. Controlling the production of urban waste in the future has become an important research topic for various environmental protection organizations. The study of China's urban waste output change rule and developing trend, not only for urban environment planning and regulatory decision-making to provide data support, but also for waste cleaning, transportation and handling to formulate reasonable implementation plan. However, the current common prediction methods still have some problems, such as low accuracy, large amount of calculation and no screening of influencing factors. Some researchers have established an RBF network forecasting model to predict the amount of urban waste in various provinces and cities across the country. The prediction of waste discharge not only has the advantages of fast convergence speed and high prediction accuracy [42].

## 3  Application status of artificial neural network in distillation

### 3.1  Distillation

Distillation is a process of separation using the different volatility of each component in the mixture. The commonly used equipment rectification tower is a typical tower type gas-liquid contact device, which can be roughly divided into a plate tower and a packed tower. The rectification tower is divided into two sections by the feed plate. Above the feed plate is called the rectification section, and below the feed plate is called the stripping section. The gas-liquid two-phase contact in the tower conducts mass and heat transfer, making the volatile components in the liquid phase enter the gas phase and come to the top of the tower, while the hardly volatile components in the gas phase are transferred to the liquid phase and come to the bottom of the tower.

Because the distillation tower plays an extremely important role in chemical applications, its implementation of soft-sensing technology has attracted much attention [43], and the artificial neural network method used in soft-sensing technology can make up for the shortcomings of traditional soft-sensing technology. Research on measuring neural network models has also attracted more and more attention. In 2008, Lu et al. [44] established two soft-sensing models based on BP neural network and fuzzy neural network respectively, and carried out dynamic identification and prediction of propylene distillation tower. The research results show that these two soft-sensing models based on artificial neural networks can well identify the dynamic behavior of the propylene distillation tower, with small errors in the prediction results and good generalization performance. In 2016, Huang [45] proposed a soft measurement method based on artificial neural network to solve the problem of material concentration measurement error at the bottom of the tower. The research established a soft-sensing model of BP neural network. The BP neural network takes the temperature of the sensitive plate of the stripping section as the input parameter and the material concentration at the bottom of the tower as the output parameter. The hidden layer has 7 nodes. The concentration at the bottom of the tower is measured. The research results show that the error of the model is not only smaller than that of the regression model, but the generalization ability of the external delay data is also stronger, which can be used to accurately detect the components of rectified products in real time.

### 3.2 Reactive distillation

Reactive rectification combines the two operations of reaction and separation together. They are carried out simultaneously in one equipment, and the products or intermediate products produced by the reaction are separated in time. This not only improves the reaction conversion rate and product yield, but also greatly reduces equipment investment and operating costs, so the reactive distillation in the industrial production has attracted much attention [46].

The use of artificial neural networks to simulate the reactive distillation process has gradually been widely studied. In 2008, Song et al. [47] used a three-layer BP neural network model to simulate the reactive distillation process of acetic acid and ethanol esterification, and on this basis, used a multi-objective genetic algorithm to optimize the reactive distillation process. The BP neural network model in this study takes the feed amount of acetic acid, the excess fraction of ethanol and the reflux ratio as the three nodes of the input layer, and the load of the condenser, the amount of ester discharged at the top of the tower, the content of ester in the top output of the tower, and the total conversion rate of acid in the feed are used as the four nodes of the output layer, and the simulation results of the neural network model are in good agreement with the results optimized by the multi-objective genetic algorithm, and the accuracy is also high, which can be applied in industrial production. In 2010, Feng et al. [48] used a three-layer BP neural network to simulate the reactive distillation process of acetic acid and methanol esterification to synthesize methyl acetate. The artificial neural network in the study adopts the momentum BP algorithm with variable learning rate and the ordinary BP algorithm respectively for training. The two BP neural network models both take the feed mole ratio of methanol and acetic acid, the feed position of methanol and the reflux ratio as the three nodes of the input layer. The heat load of the tower top condenser, the heat load of the tower kettle reboiler, the content of methyl acetate in the tower top discharge and the conversion rate of acetic acid in the feed are taken as the four nodes of the output layer, and the optimal number of nodes in the hidden layer is 14 by comparison. The research results show that the prediction error of the artificial neural network under the training of two different BP algorithms is not large, but the neural network trained with the momentum BP algorithm with variable learning rate has higher simulation prediction results. In the same year, Lang et al. [49] used the RBF artificial neural network to simulate the reactive distillation process of the esterification of acetic acid with ethanol to synthesize ethyl acetate, and on this basis, used the ant colony immune algorithm to optimize. The input parameters of the RBF neural network model used in this study are the feed amount of acetic acid, the mass fraction of ethanol and the reflux ratio, and the output parameters are the load of the condenser, the amount of ester discharged from the top of the tower, the amount of ester discharged from the top of the tower, and the total conversion rate of acid in the feed. The simulation results of this model are in good agreement with the results optimized by the ant colony immune algorithm, indicating the accuracy of the RBF neural network model and the high accuracy of the optimized results of the ant colony immune algorithm.

### 3.3 Catalytic distillation

Catalytic distillation combines the catalytic reaction with the rectification process. It is realized by filling the solid catalyst in the rectification tower in an appropriate form. This rectification method has the advantages of high selectivity, high yield, and energy saving, so it is used increasingly wide [50].

The application of artificial neural networks in the research and development of catalytic distillation technology has also become more common. As early as 1998, Xiao et al. [51] used artificial neural networks to simulate and predict the process of hydrolysis of methyl acetate in catalytic distillation towers. The research used BP algorithm and simulated annealing algorithm to train the artificial neural network, and the input parameters are the water-to-ester ratio of the feed, the reflux feed ratio, and the volume flow rate of methyl acetate to the catalyst volume ratio, and the output parameters are the conversion rate of methyl acetate and the acid-water ratio in the bottom of the tower. The research results show that the prediction results of the artificial neural network model are very consistent with the experimental results. The artificial neural network model can be used to simulate and predict the catalytic distillation operation, and it can be used to optimize and improve the catalytic distillation process conditions. In 2006, Hu et al. [52] also used the artificial neural network to simulate the start-up process of the catalytic distillation tower. The learning algorithm of the artificial neural network used in this study is the L-M method, and the input parameter is the initial liquid of the tower. The output parameter is the time taken to start the catalytic distillation. The prediction performance of the artificial neural network increases with the increase of training data, and the prediction error gradually decreases, indicating that the artificial neural network can be used to predict during the start-up of the catalytic distillation tower.

### 3.4 Thermally Coupled Distillation

Thermally coupled rectification is mainly used for the separation of three-component and above mixtures. It is an energy-saving rectification, which conforms to the development trend of the current era, and has attracted widespread attention [53].

The traditional thermally coupled distillation simulation calculation process is very complicated, and the artificial neural network makes up for the shortcomings of the traditional method. In 2003, Wang et al. [54] used artificial neural network to simulate the thermally coupled distillation process and optimized it with genetic algorithm. The artificial neural network in this study is trained using an improved BP algorithm. The input layer includes four nodes: the top temperature of the auxiliary tower, the top temperature of the main tower, the liquid phase reflux flow rate of the main tower, and the vapor phase reflux flow rate of the main tower. The output layer includes three nodes: the production volume of main tower product B, the production volume of main tower product C, and the heat load of the reboiler. And the hidden layer has four nodes. The research results show that the simulation calculation results are in good agreement with the prediction results and optimization results, and the accuracy is high, indicating that the neural network can be used to simulate the thermally coupled distillation process. In the same year, He et al. [55] also used artificial neural networks and genetic algorithms in the study of butadiene separation and acetonitrile recovery processes. This study also used a three-layer BP neural network. The input parameters are raw material temperature, mass separator feed temperature, main tower reflux ratio, main tower gas recovery volume and mass separator circulation volume. And the output parameters are the heat load of the reboiler, the alkyne content of the butadiene in the top distillate of the main tower, and the alkyne content of the mass separator discharged from the bottom of the tower. The simulation results of the neural network are basically consistent with the predicted and optimized results, indicating that the artificial neural network can better simulate the thermally coupled distillation process.

## 3.5 Decompressing rectify

Vacuum rectification is a rectification operation that separates the mixture under reduced pressure, and can be used to separate some substances that are easy to decompose or polymerize during rectification at high temperatures.

In the research of vacuum distillation, there are also applications of artificial neural networks. In 2012, Chen et al. [56] used an artificial neural network to simulate the separation of stable isotope $^{18}$O by vacuum distillation. The artificial neural network was trained using an improved BP algorithm. The input parameters were product recovery and tower top pressure. The output parameter is the $^{18}$O abundance of the product, and the hidden layer has four nodes. The results show that the prediction results are basically consistent with the simulation calculation results, and the accuracy is high. Therefore, the artificial neural network can be used to simulate and predict the vacuum distillation process.

## 4 Concluding

In summary, artificial neural networks have been widely used in distillation. Its advantages such as self-learning, associative storage, and high-speed search for optimized solutions make up for some of the shortcomings of traditional methods, and make it more concerned in the research field. With the development of science and technology, artificial neural networks will inevitably become more mature, and their applications in various chemical processes will also become more common.